\documentclass[10pt,twocolumn,letterpaper]{article}

\usepackage{iccv}
\usepackage{times}
\usepackage{epsfig}
\usepackage{graphicx}
\usepackage{amsmath}
\usepackage{amssymb}
\usepackage{algpseudocode}
\usepackage{algorithm}

\usepackage[pagebackref=true,breaklinks=true,letterpaper=true,colorlinks,bookmarks=false]{hyperref}

\iccvfinalcopy 

\def\iccvPaperID{****} 

\ificcvfinal\fi
\begin{document}

\title{SC-FEGAN: Face Editing Generative Adversarial Network with User's Sketch and Color}

\author{Youngjoo Jo\qquad Jongyoul Park\\ 
ETRI\\
South Korea\\
{\tt\small \{run.youngjoo,jongyoul\}@etri.re.kr}
}

\maketitle

\begin{abstract}
   We present a novel image editing system that generates images as the user provides free-form mask, sketch and color as an input.
   Our system consist of a end-to-end trainable convolutional network.
   Contrary to the existing methods, our system wholly utilizes free-form user input with color and shape.
   This allows the system to respond to the user's sketch and color input, using it as a guideline to generate an image.
   In our particular work, we trained network with additional style loss~\cite{gatys2016image} which made it possible to generate realistic results, despite large portions of the image being removed.
   Our proposed network architecture SC-FEGAN is well suited to generate high quality synthetic image using intuitive user inputs.
\end{abstract}

\begin{figure}[t]
\begin{center}
   \includegraphics[width=1.0\linewidth]{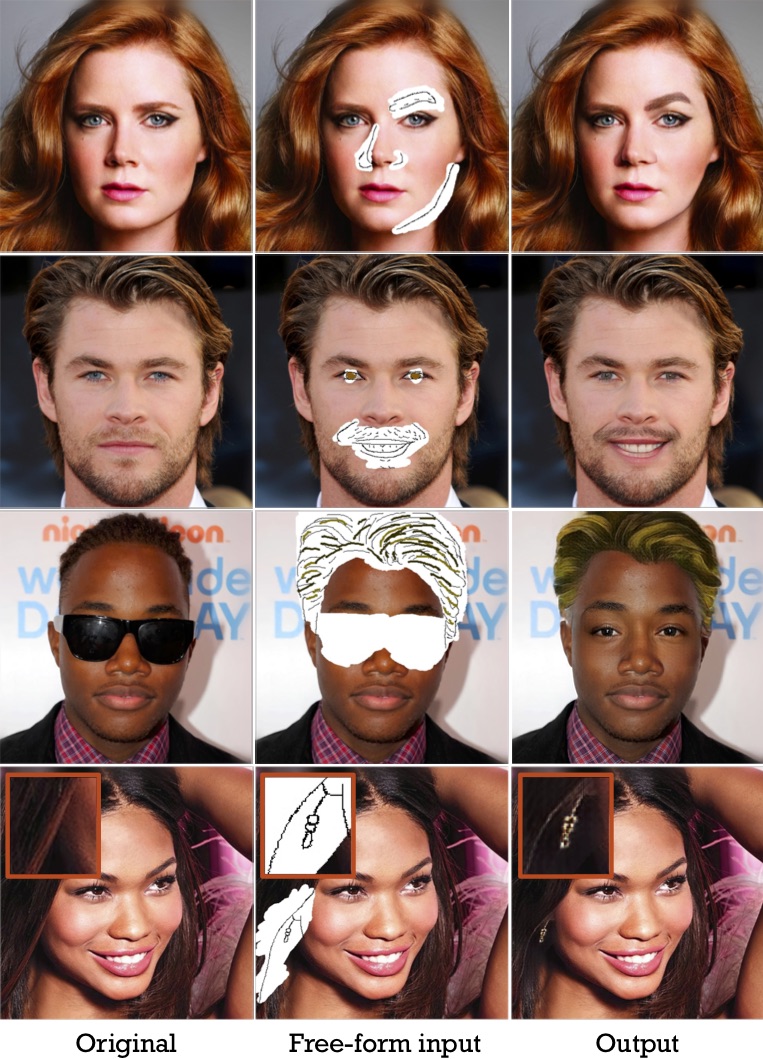}
\end{center}
   \caption{Face image editing results by our system. It can takes free-form input consist of mask, sketch and color. For each example, it shows that our system make users can easily edit shape and color of face even if user wants completely change hairstyle and eye (third row). Interestingly, user can edits earring by our system (fourth row).}
\label{fig:1}
\end{figure}

\section{Introduction}

Image completion with generative adversarial networks (GANs) is a highly recognized topic in computer vision.
With image exchange becoming a common medium in daily communication in the present day, there is an increase of demand for realism in generated image over a minimal image completion feature.
Such demand is reflected on social media statistics.
However, most image editing softwares require expertise such as knowing which specific tools to use at certain circumstances to effectively modify the image the way we want.
Instead, an image completion method which responds to user input would allow novice to easily modify images as desired.
Similarly, our proposed system has an ability to easily produce high quality face images, provided a sketch and color input is given even with the presence of erased parts in the image.

In recent works, deep learning based image completion methods have been used to restore erased portion of an image.
The most typical method used an ordinary (square) mask, then restored the masked region with an encoder-decoder generator.
Then global and local discriminator was used to make an estimation on whether the result was real or fake~\cite{iizuka2017globally,li2017generative}.
However, this system is limited to low resolution images, and the generated image had awkward edges on of the masked regions.
In addition, the synthesized images on the restored region often fell short of the user’s expectation as the generator was never given any user input to utilize as a guide.
Several works that improved on this limitation include Deepfillv2~\cite{yu2018free}, a work that utilized user's sketch as an input, and GuidedInpating~\cite{zhao2018guided}, which took a part of another image as an input to restore the missing parts.
However, since Deepfillv2 does not use color input, the color in the synthesized image is gerated by inference from the prior distribution learned from the training data set.
The GuidedInpating used parts of other images to restore deleted regions.
However it was difficult to restore in detail because such process required inferring the user's preferred reference image.
Another recent work Ideepcolor~\cite{zhang2017real} proposed a system that accepts color of user input as reference to create a color image for black and white images.
However, the system in Ideepcolor does not allow editing of the object structures or restore deleted parts on image.
In another work, a face editing system FaceShop~\cite{portenier2018faceshop} which accepts sketch and color as user input was introduced.
However, FaceShop has some limitations to be used as an interactive system for synthetic image generation.
Firstly, it utilized random rectangular rotable masks to erase the regions that are used in local and global discriminator.
This means that local discriminator must resize the restored local patch to accept fitting input dimensions, and the resizing process would distort the information in both the erased and remaining portions of the image.
As a result, the produced image would have awkward edges on the restored portion.
Secondly, FaceShop would produce an unreasonable synthetic image if too much area is wiped out.
Typically, when given an image with the entire hair image erased, system restores it with distorted shape.

To deal with the aforementioned limitations, we propose a SC-FEGAN with a fully-convolutional network that is capable of end-to-end training.
Our proposed network uses a SN-patchGAN~\cite{yu2018free} discriminator to address and improve on the awkward edges.
This system is trained with not only general GAN loss but also concurrently with style loss to edit the parts of the face image even if a large area is missing.
Our system creates high quality realistic composite images with the user's free-form input.
The free-form domain input of sketch and color also has an interesting additive effects, as shown in Figure~\ref{fig:1}.
In summary, we make the following contributions:
\begin{quotation}
\noindent
\noindent\textbullet \; We suggest a network architecture similar to Unet~\cite{ronneberger2015u} with gated convolutional layers~\cite{yu2018free}.
Such architecture is easier and faster for both training and inference stages.
It produced superior and detailed result compared to the Coarse-Refined network in our case.\\

\noindent\textbullet \; We created a free-form domain data of masks, color and sketch.
This data is used for making incomplete image data for training instead of stereotyped form input.\\

\noindent\textbullet \;We applied SN-patchGAN~\cite{yu2018free} discriminator and trained our network with additional style loss.
This application covers cases with large portions erased and has shown robustness at managing the edges of the masks.
It also allowed production of details on the produced image such as high quality synthetic hair style and earring.\\
\end{quotation}

\section{Related Work}
Interactive image modification has an extensive history, predominantly in regards with the techniques that use hand-crafted features rather than deep learning techniques.
Such predominance is reflected in commercial image editing software and our practice of its usage.
Because most commercial image editing softwares use defined operations, a typical image modification task requires expert knowledge to strategically apply a combination of transformations for an image.
In addition to expert knowledge, users are required to devote long working hours to produce a delicate product.
Therefore traditional approach is disadvantageous for the non-experts and is tedious to use for producing high quality results.
In addition to these conventional modeling methods, recent breakthroughs in GAN research have developed several methods for completion, modification, and transformation of images by training generative models with large data sets.

In this section, we discuss several works in the field of image completion and image translation among prevalent image editing methods that use deep learning.

\subsection{Image Translation}
GAN for image translation was first proposed for learning image domain transforms between two datasets~\cite{CycleGAN2017, pix2pix2017}.
Pix2Pix~\cite{pix2pix2017} proposed a system used dataset which consists of pair of images that can be used to create models that convert segmentation labels to the original image, or convert a sketch to an image, or a black and white image to color image.
But this system requires that images and target images must exist as a pair in the training dataset in order to learn the transform between domains.
CycleGAN~\cite{CycleGAN2017} suggests an improvement over such requirement.
Given a target domain without a target image, there exists a virtual result in the target domain when converting the image in original domain.
If the virtual result is inverted again, the inverted result must be the original image. So, it takes two generators for converting task.

Recently, following the domain to domain change, several researches efforts have demonstrated systems that can take user input for adding the desired directions to generated results.
The StarGAN~\cite{StarGAN2018} used a single generator and a discriminator to flexibly translate an input image to any desired target domain by training with domain label.
The Ideepcolor~\cite{zhang2017real} is introduced as a system that convert a monochrome image into a color image by taking a user's desired color as a mask.
In these works, image transformation that interacts with user input has shown that user input can be learned by feeding it to the generator with images.

\subsection{Image Completion}
Image completion field has two main challenges: 1) Filling deleted area of the image, 2) proper reflection of users input in the restored area.
In a previous study, a GAN system explores the possibility of generating complete images with erased area~\cite{iizuka2017globally}.
It uses generator from the U-net~\cite{ronneberger2015u} structure and utilize local and global discriminator.
The discriminator decides whether the generated image is real or fake on both the newly filled parts and full reconstructed image respectively.
The Deepfillv1~\cite{yu2018generative} also used rectangle mask and global and local discriminator model to suggest that contextual attention layer extensively improves the performance.
However, the global and local discriminator still produces awkward region at the border of restored parts.

In the follow-up study Deepfillv2~\cite{yu2018free}, a free-form mask and SN-patchGAN were introduced to replace the existing rectangular mask and global and local discriminator with single discriminator.
In addition, the gated convolutional layer that learns the features of the masked region was also suggested.
This layer can present masks automatically from data by training, which gives
the network ability to reflect user sketch input on the results.

Our proposed network described in the next section allows the usage of not only sketch but also color data as an input for editing the image.
Even though we utilized a U-net structure instead of a Coarse-Refined net structure such as in Deepfillv1,2~\cite{iizuka2017globally,yu2018free}, our network generate high-quality results without complex training schedule nor with requirement of other complex layers.

\section{Approach}
In this paper, we describe the proposed SC-FEGAN, a neural network based face image editing system and also describe methods for making the input batch data.
This network can be trained end-to-end and generates high quality synthetic image with realistic texture details.

In Section 3.1, we discuss our method for making training data.
In Section 3.2, we describe our network structure and loss functions that allow extraction of features from sketch and color input, while simultaneously achieving stability in training.

\subsection{Training Data}
Suitable training data is a very important factor for increasing the training performance of the network and increasing the responsiveness to user input.
To train our model, we used the CelebA-HQ~\cite{karras2017progressive} dataset after several pre-processing steps as described as following.
We first randomly select 2 sets of 29,000 images for training and 1,000 images for testing.
We resize the images to 512$\times$512 pixels before attaining the sketch and color dataset.

To better express the complexity of the eye in the face image, we used a free-from mask based on the eye position to train network.
Also, we created appropriate sketch domains and color domains by using free-form mask and face segmentation GFC~\cite{li2017generative}.
This was a crucial step which enabled our system to produce persuasive results for the hand-drawn user input case.
We randomly applied masks to hair region in input data because it has different properties compared other parts of face.
We discuss more details below.

{\bf Free-form masking with eye-positions} We used a similar masking method as that which was presented in Deepfillv2~\cite{yu2018free} to make incomplete images.
However, when training on the facial images, we randomly applied a free draw mask with eye-positions as a starting point in order to express complex parts of the eyes.
We also randomly added hair mask using GFC~\cite{li2017generative}.
Details are described in Algorithm 1.

\begin{figure}[t]
\begin{center}
   \includegraphics[width=1.0\linewidth]{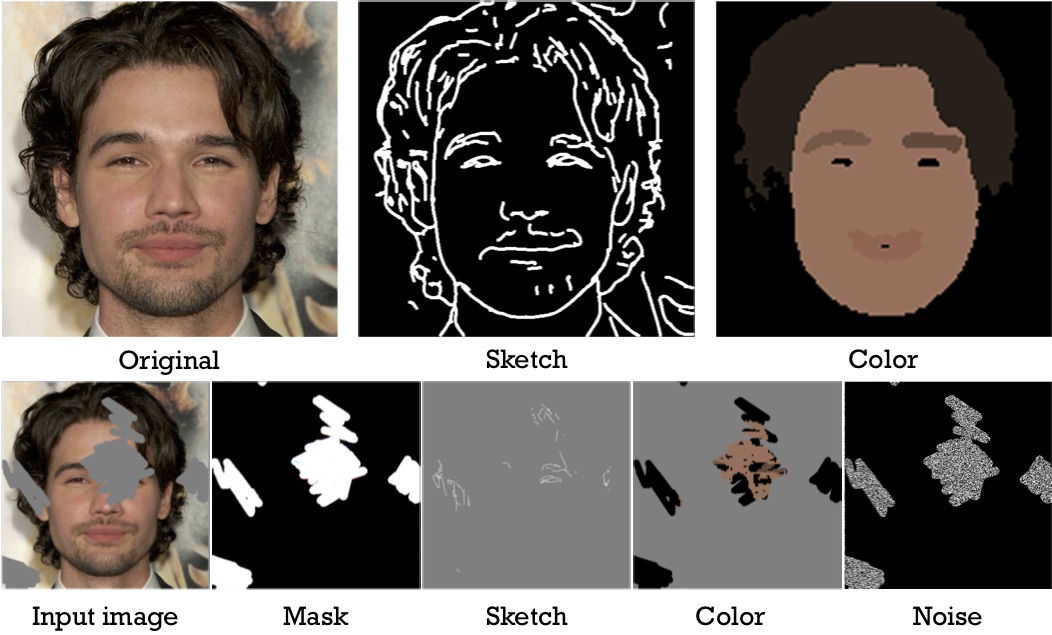}
\end{center}
   \caption{Sketch and color domain dataset and inputs for batch. We extract sketches using HED edge detector~\cite{xie15hed}. The color maps are generated by median color of segmented areas from using GFC~\cite{li2017generative}. The inputs of network consist of incomplete image, mask, sketch, color and noise.}
\label{fig:2}
\end{figure}

\begin{algorithm}
\caption{Free-form masking with eye-positions}
\begin{algorithmic}
\State maxDraw, maxLine, maxAngle, maxLength are hyper parameters
\State GFCHair is the GFC for get hair mask of input image
\State Mask=zeros(inputSize,inputSize)
\State HairMask=GFCHair(IntputImage)
\State numLine=random.range(maxDraw)
\For {$i$=0 to numLine}
\State startX = random.range(inputSize)
\State startY = random.range(inputSize)
\State startAngle = random.range(360)
\State numV = random.range(maxLine)
\For {$j$=0 to numV}
\State angleP = random.range(-maxAngle,maxAngle)
\If {$j$ is even}
\State angle = startAngle+angleP
\Else
\State angle = startAngle+angleP+180
\EndIf
\State length = random.range(maxLength)
\State Draw a line on Mask from point (startX, startY) with angle and length.
\State startX = startX + length * sin(angle)
\State startY = stateY + length * cos(angle)
\EndFor
\State Draw a line on Mask from eye postion randomly.
\EndFor
\State Mask = Mask + HairMask (randomly)
\end{algorithmic}
\end{algorithm}

\indent {\bf Sketch} \& {\bf Color domain} For this part, we used a method similar to that used in FaceShop~\cite{portenier2018faceshop}.
However, we excluded AutoTrace~\cite{autotrace} which converts bitmap to vector graphics for sketch data.
We used the HED~\cite{xie15hed} edge detector to generate the sketch data whcih corresponds to the user's input to modify the facial image.
After that, we smoothed the curves and erased the small edges.
To create color domain data, we first created blurred images by applying a median filtering with size 3 followed by 20 application of bilateral filter.
After that, GFC~\cite{li2017generative} was used to segment the face, and each segmented parts were replaced with the median color of the corresponding parts.
When creating data for the color domain, histogram equalization was not applied to avoid color contamination from light reflection and shadowing.
However, because it was more resonant for users to express all part of face in sketch domain regardless of blur caused by interference of light, histogram equalization was used when creating the sketch from the domain data.
More specifically, after histogram equalization, we applied HED to get the edges from the images. Then, we smoothed the curve and erased the small objects.
Finally, we multiplied the mask, adopting a process similar to the previous free-form mask, and color images and get color brushed images.
See Figure~\ref{fig:2} for an example of our data.

\subsection{Network Architecture}
Inspired by recent image completion studies~\cite{iizuka2017globally, yu2018free, portenier2018faceshop}, our completion network (\textit{i.e.} Generator) is based on encoder-decoder architecture like the U-net~\cite{ronneberger2015u} and our discrimination network is based on SN-patchGAN~\cite{yu2018free}.
Our network structure produces high-quality synthesis results with image size of 512$\times$512 while achie ving stable and fast training.
Our network also trains generator and discriminator simultaneously like the other networks.
The generator receives incomplete images with user input to create an output image in the RGB channel, and inserts the masked area of the output image into the incomplete input image to create a complete image.
The discriminator receives either a completed image or an original image (without masking) to determine whether the given input is real or fake.
In adversarial training, additional user input to the discriminator also helps to improve performance.
Also, we found that additional loss that is different to general GAN loss is effective to restore large erased portions. 
Details of our network are shown below.

{\bf Generator} Figure~\ref{fig:3} shows our network architecture in detail.
Our generator is based on U-net~\cite{liu2018image} and all convolution layers use gated convolution~\cite{yu2018free} using 3x3 size kernel.
Local signal normalization (LRN)~\cite{karras2017progressive} is applied after feature map convolution layers excluding other soft gates.
LRN is applied to all convolution layers except input and output layers.
The encoder of our generator receives input tensor of size 512$\times$512$\times$9: an incomplete RGB channel image with a removed region to be edited, a binary sketch that describes the structure of removed parts, a RGB color stroke map, a binary mask and a noise (see Figure~\ref{fig:2}).
The encoder downsamples input 7 times using 2 stride kernel convolutions, followed by dilated convolutions before upsampling.
The decoder uses transposed convolutions for upsampling.
Then, skip connections were added to allow concatenation with previous layer with the same spatial resolution.
We used the leaky ReLU activation function after each layer except for the output layer, which uses a tanh function.
Overall, our generator consists of 16 convolution layers and the output of the network is an RGB image of same size of input (512$\times$512).
We replaced the remaining parts of image outside the mask with the input image before applying the loss functions to it.
This replacement allows the generator to be trained on the edited region exclusively.
Our generator is trained with losses which were introduced in PartialConv~\cite{liu2018image}: per-pixel losses, perceptual loss, style loss and total variance loss.
The generic GAN loss function is also used.

{\bf Discriminator} Our discriminator has SN-PatchGAN~\cite{yu2018free} structure.
Unlike Deepfillv2~\cite{yu2018free}, we did not apply ReLu function on the GAN loss.
Also we used 3$\times$3 size convolution kernel and applied gradient penalty loss term.
We added an extra term, to avoid the discriminator output patch reaching value close to zero.
Our overall loss functions are shown as below:

\begin{figure*}[t]
\begin{center}
   \includegraphics[width=1.0\linewidth]{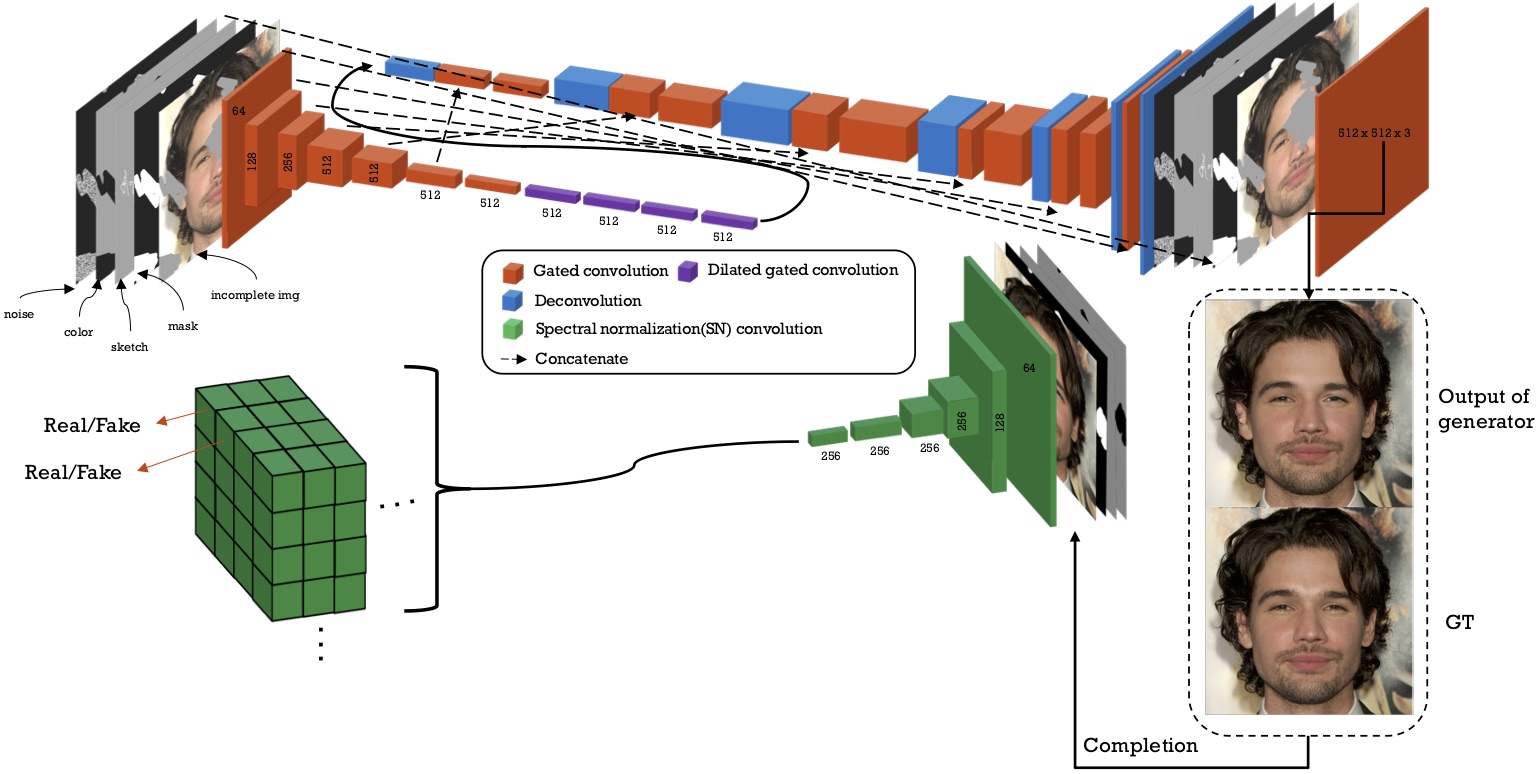}
\end{center}
   \caption{Network architecture of SC-FEGAN. LRN is applied after all convolution layers except input and output. We used tanh as activation function for output of generator. We used SN convolution layer~\cite{miyato2018spectral} for discriminator.}
\label{fig:3}
\end{figure*}
\begin{align}
{ L }_{ G\_ SN }=-{\rm I\!E}\left[ D\left( { I }_{ comp } \right)  \right],
\end{align}
\begin{align}
{ L }_{G}={ L }_{ per-pixel }+\sigma{ L }_{ percept }+\beta{ L }_{ G\_SN }\\\notag+ \gamma({ L }_{ sytle }\left( { I }_{ gen } \right)+{ L }_{ sytle }\left( { I }_{ comp } \right))\\\notag+\upsilon{ L }_{ tv }+\epsilon{\rm I\! E}\left[ {D({ { I }_{ gt } })}^{2} \right],
\end{align}
\begin{align}
{ L }_{ D }={\rm I\! E}\left[ 1 - D({ I }_{ gt }) \right]
{}+{\rm I\! E}\left[ 1 + D({ I }_{ comp }) \right]+\theta{ L }_{ GP }.
\end{align}

Our generator was trained with ${L}_{G}$ and discriminator was trained with ${L}_{D}$.
$D(I)$ is the output of the discriminator given input $I$.
Additional losses, ${L}_{sytle}$ and ${L}_{percept}$ are critical when editing large region such as hairstyle.
The details of each loss are described below.
The ${L}_{per-pixel}$ of $L^{1}$ distances between ground truth image ${I}_{gt}$ and the output of generator ${I}_{gen}$ is computed as
\begin{flalign}
{ L }_{ per-pixel }=\frac { 1 }{ { N_{{ I }_{ gt }} } } \left\| M\odot \left( { I }_{ gen }-{ I }_{ gt } \right)  \right\|_{1}
\\\notag +\alpha \frac { 1 }{ N_{{ I }_{ gt }} } \left\| \left( 1-M \right) \odot \left( { I }_{ gen }-{ I }_{ gt } \right)  \right\|_{1},
\end{flalign}
where, $N_{a}$ is the number elements of feature $a$, $M$ is the binary mask map and ${I}_{gen}$ is the output of generator.
We used the factor $\alpha>1$ to give more weight the loss on the erased part.
The perceptual loss, ${L}_{percept}$, also computes $L^{1}$ distances, but after projecting images into feature spaces using VGG-16~\cite{russakovsky2015imagenet} which are pre-trained on ImageNet.
It is computed as
\begin{flalign}
{ L }_{ percept }=\sum _{ q }{ \frac { \left\| { \Theta  }_{q}\left( { I }_{ gen }\right) -{ \Theta  }_{q}\left( { I }_{ gt }\right)  \right\| _{ 1 } }{ { N }_{ { \Theta  }_{q}\left( { I }_{ gt }\right)}}+}\\\notag\sum _{ q }{ \frac { \left\| { \Theta  }_{q}\left( { I }_{ comp }\right) -{ \Theta  }_{q}\left( { I }_{ gt }\right)  \right\| _{ 1 } }{ { N }_{ { \Theta  }_{q}\left( { I }_{ gt }\right)  } }  }.
\end{flalign}
Here, ${\Theta}_{q}(x)$ is the feature map of the $q$-th layer of VGG-16~\cite{russakovsky2015imagenet} given that input $x$ is given and ${I}_{comp}$ is the completion image of ${I}_{gen}$ with the non-erased parts directly
set to the ground truth.
$q$ is the selected layers from VGG-16, and we used layers of $pool1,pool2$ and $pool3$.
Style loss compare the content of two images by using Gram matrix. We compute the style loss as
\begin{align}
{ L }_{ sytle }\left( I \right) =\sum _{ q }{ \frac { 1 }{ { C }_{ q }{ C }_{ q } } \left\| \frac { \left( {G}_{q}\left( I\right) -{G}_{q}\left( { I }_{ gt }\right)  \right) }{ { N }_{ q } }   \right\| _{ 1 } },
\end{align}
where the ${G}_{q}(x)=({\Theta}_{q}(x))^{T}({\Theta}-{q}(x))$ is the Gram matrix to perform an autocorrelation on each feature maps of VGG-16.
When the feature is of shape ${H}_{q}\times{W}_{q}\times{C}_{q}$, the output of Gram matrix is of shape ${C}_{q}\times{C}_{q}$.

${ L }_{ tv } = { L }_{ tv-col }+{ L }_{ tv-row }$ is total variation loss suggested by fast-neural-style~\cite{johnson2016perceptual} to improve the \textit{checkerboard artifacts} from perceptual loss term.
It is computed as
\begin{align}
{ L }_{ tv-col }=\sum _{ (i,j)\in R}^{  }{ \frac { \left\| { I }_{ comp }^{ i,j+1 }-{ I }_{ comp }^{ i,j } \right\| _{ 1 } }{ N_{ comp } }  },
\end{align}

\begin{align}
{ L }_{ tv-row }=\sum _{ (i,j)\in R }^{  }{ \frac { \left\| { I }_{ comp }^{ i+1,j }-{ I }_{ comp }^{ i,j } \right\| _{ 1 } }{ N_{ comp } }  },
\end{align}
where $R$ is the region of the erased parts.
The WGAN-GP~\cite{gulrajani2017improved} loss is used for improving training and is computed as
\begin{align}
{ L }_{ GP } =    
{\rm I\! E} \left[ { \left( { \left\| { \nabla  }_{ U }D(U)\odot M \right\|  }_{ 2 }-1 \right)  }^{ 2 } \right].
\end{align}
Here, $\mathbf{U}$ is a data point uniformly sampled along the straight line between discriminator inputs from ${I}_{comp}$ and ${I}_{gt}$.
This term is critical to quality of synthetic image in our case.
We used $\sigma=0.05, \beta=0.001, \gamma=120, \upsilon=0.1, \epsilon=0.001$ and $\theta=10$.

\section{Results}
In this section, we present ablation studies with comparisons to recent related works, followed by face editing results. All experiments were executed on NVIDIA(R) Tesla(R) V100 GPU and Power9 @ 2.3GHz CPU with Tensorflow~\cite{abadi2016tensorflow} v1.12, CUDA v10, Cudnn v7 and Python 3. For testing, it takes \textbf{44ms} on GPU, \textbf{53ms} on CPU for resolution 512$\times$512 on average regardless of size and shape of inputs.
The source code and more results are displayed in \url{https://github.com/JoYoungjoo/SC-FEGAN}.
\begin{figure}[t]
\begin{center}
   \includegraphics[width=1.0\linewidth]{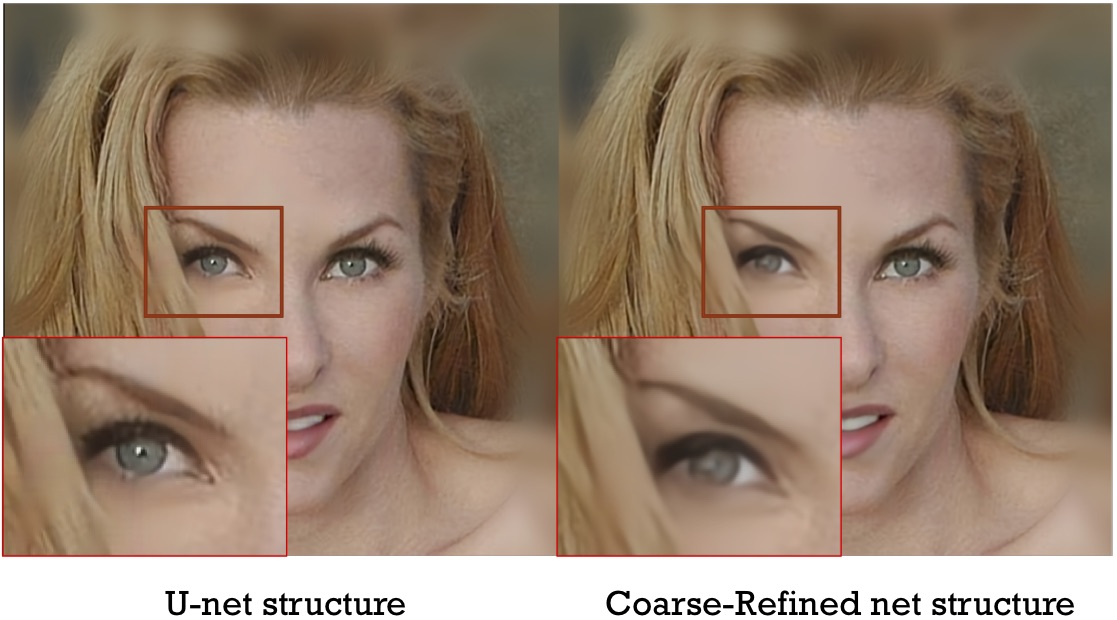}
\end{center}
   \caption{Our results with U-net(Left) and Coarse-Refined net(Right) when eye regions are removed.}
\label{fig:4}
\end{figure}
\begin{figure}
\begin{center}
   \includegraphics[width=1.0\linewidth]{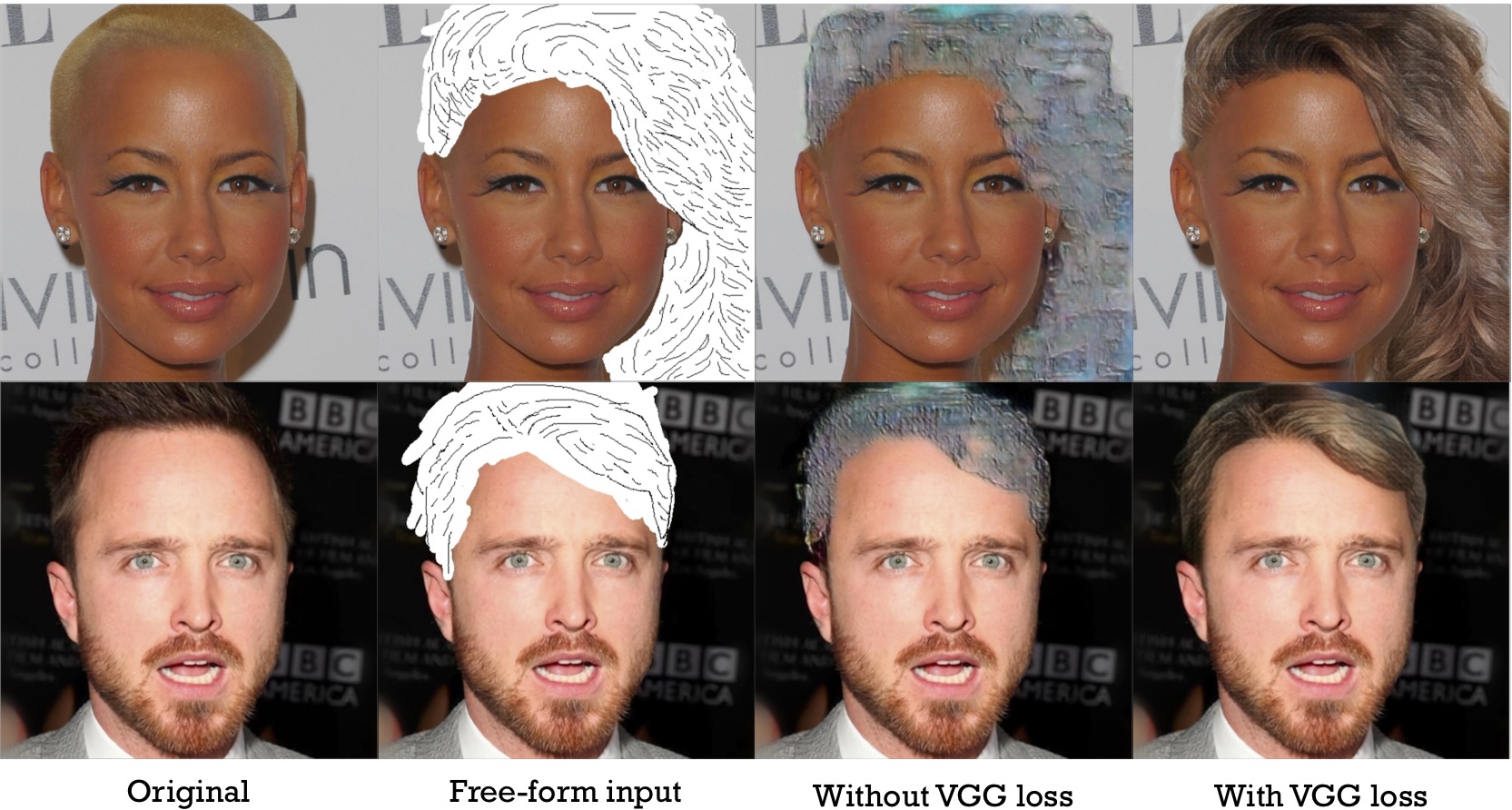}
\end{center}
   \caption{Our results from trained networks with and without VGG loss. When the network is trained without VGG loss, we encountered the similar problems as the FaceShop~\cite{portenier2018faceshop}.}
\label{fig:5}
\end{figure}
\begin{figure}
\begin{center}
   \includegraphics[width=1.0\linewidth]{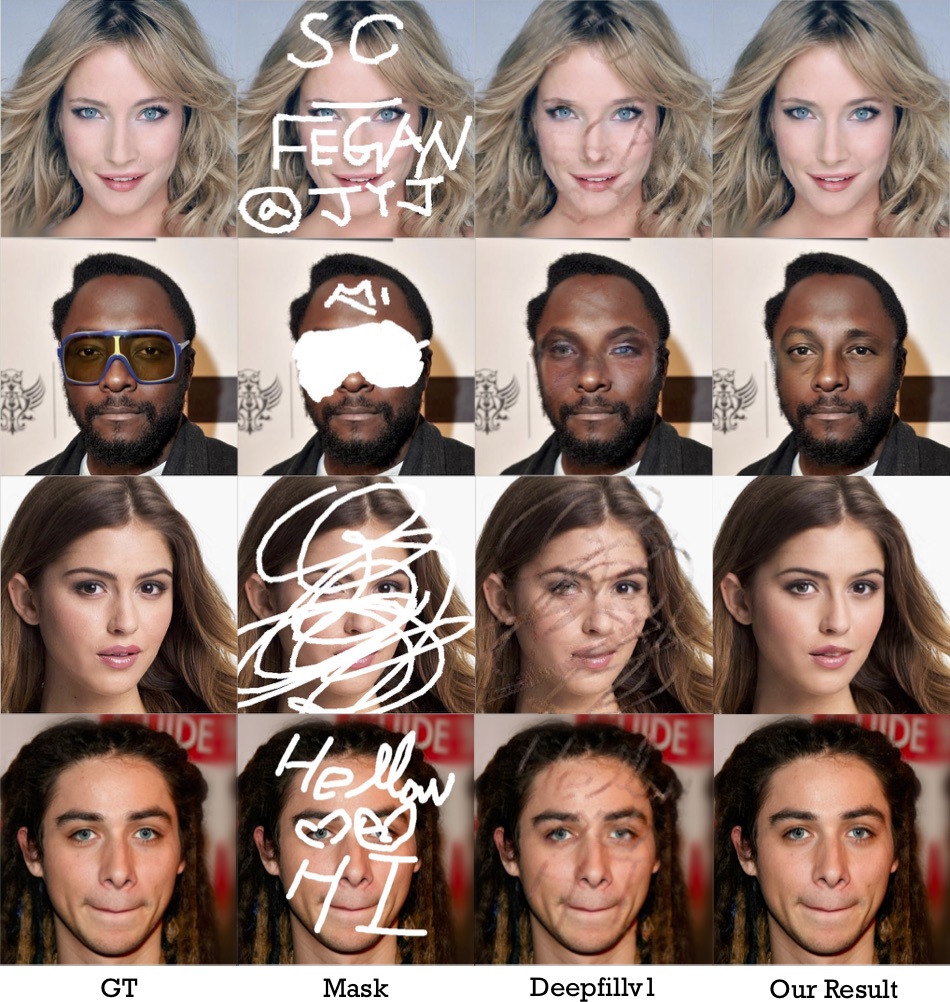}
\end{center}
   \caption{Qualitative comparisons with Deepfillv1~\cite{yu2018generative} on the CelebA-HQ validation sets.}
\label{fig:6}
\end{figure}
\begin{figure*}
\begin{center}
   \includegraphics[width=1.0\linewidth]{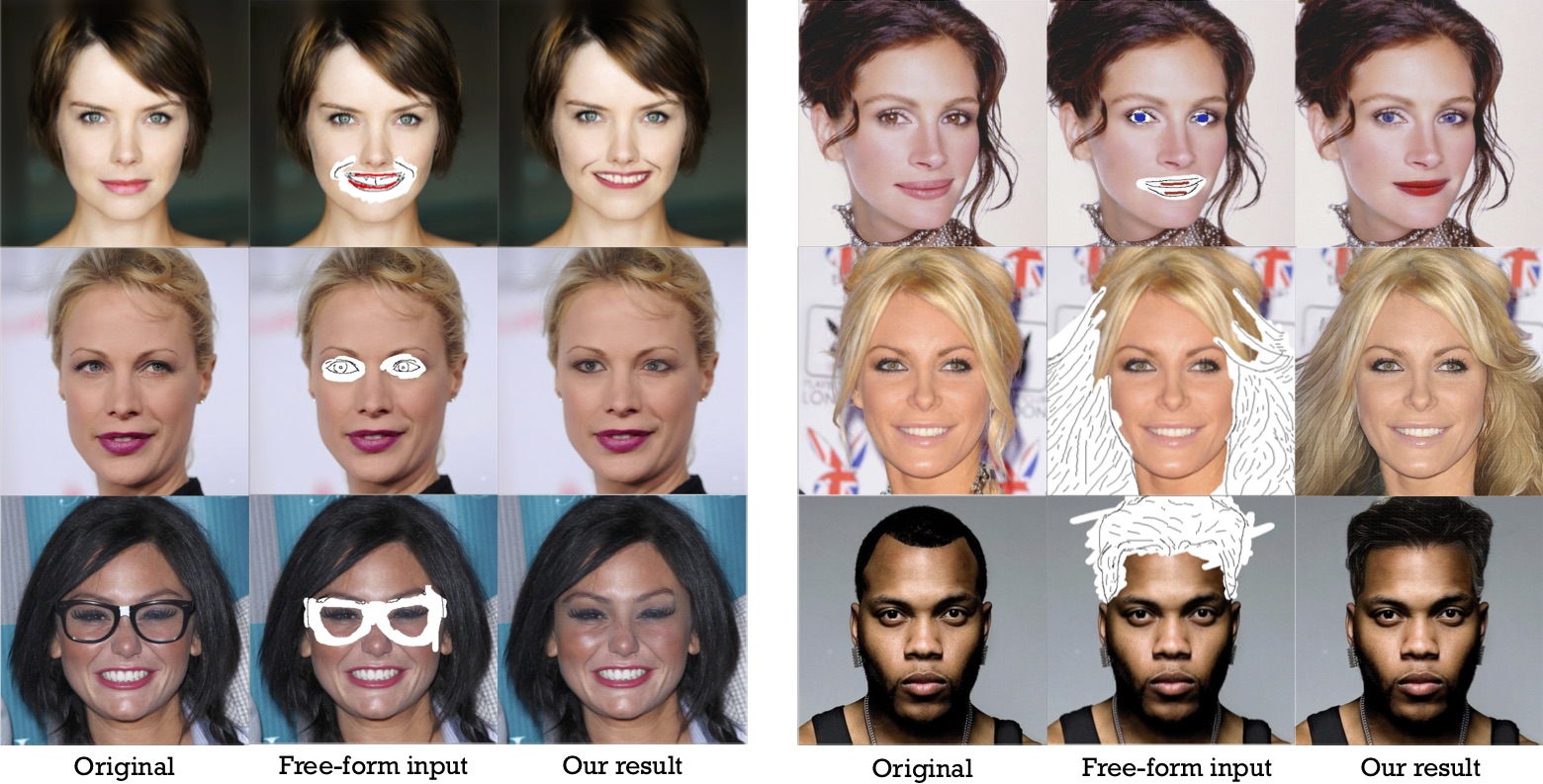}
\end{center}
   \caption{Face image editing results from our system. It shows that our system can change the shape and color of face properly. It also shows that it can be used in changing the color of eyes or erasing unnecessary parts. In particular, the right bottom two results show that our system can also be used for new hairstyle modification.}
\label{fig:7}
\end{figure*}
\subsection{Ablation Studies and Comparisons}
We first compared our results to Coarse-Refined structure network and U-net structure network.
In Deepfillv2~\cite{yu2018free}, it presents that Coarse-Refined structure and contextual attention module is effective for generation.
However, we tested Coarse-Refined structure network and notice that the refining stage has blurred the output.
We discovered that the reason for this is because $L^{1}$ loss about output of refined network is always smaller than of coarse network.
The coarse network generates a coarse estimate of the recovered region by using incomplete input.
This coarse image is then passed to the refined network.
Such setup allows refined network to learn the transform between ground truth and the coarsely recovered about incomplete input.
To achieve such effect with convolution operation, blurring on the input data is used as a workaround for an otherwise much complicated training method.
It can ameliorate the \textit{checkerboard artifacts} but it needs a lot of memories and time to training.
Figure~\ref{fig:4} shows the result of our system about Coarse-Refined structure network.

The system in FaceShop~\cite{portenier2018faceshop} has shown difficulty in modifying the huge erased image like whole hair regions.
Our system performs better in that regard due to perceptual and style losses.
Figure~\ref{fig:5} shows the result of with and without VGG loss.
We also conducted a comparison with the recent research Deepfillv1~\cite{yu2018generative} in which the test system was published.
Figure~\ref{fig:6} shows that our system produces better results in terms of quality for structure and shape with free-form masks.
\subsection{Face Image Editing and Restoration}
\begin{figure*}
\begin{center}
   \includegraphics[width=1.0\linewidth]{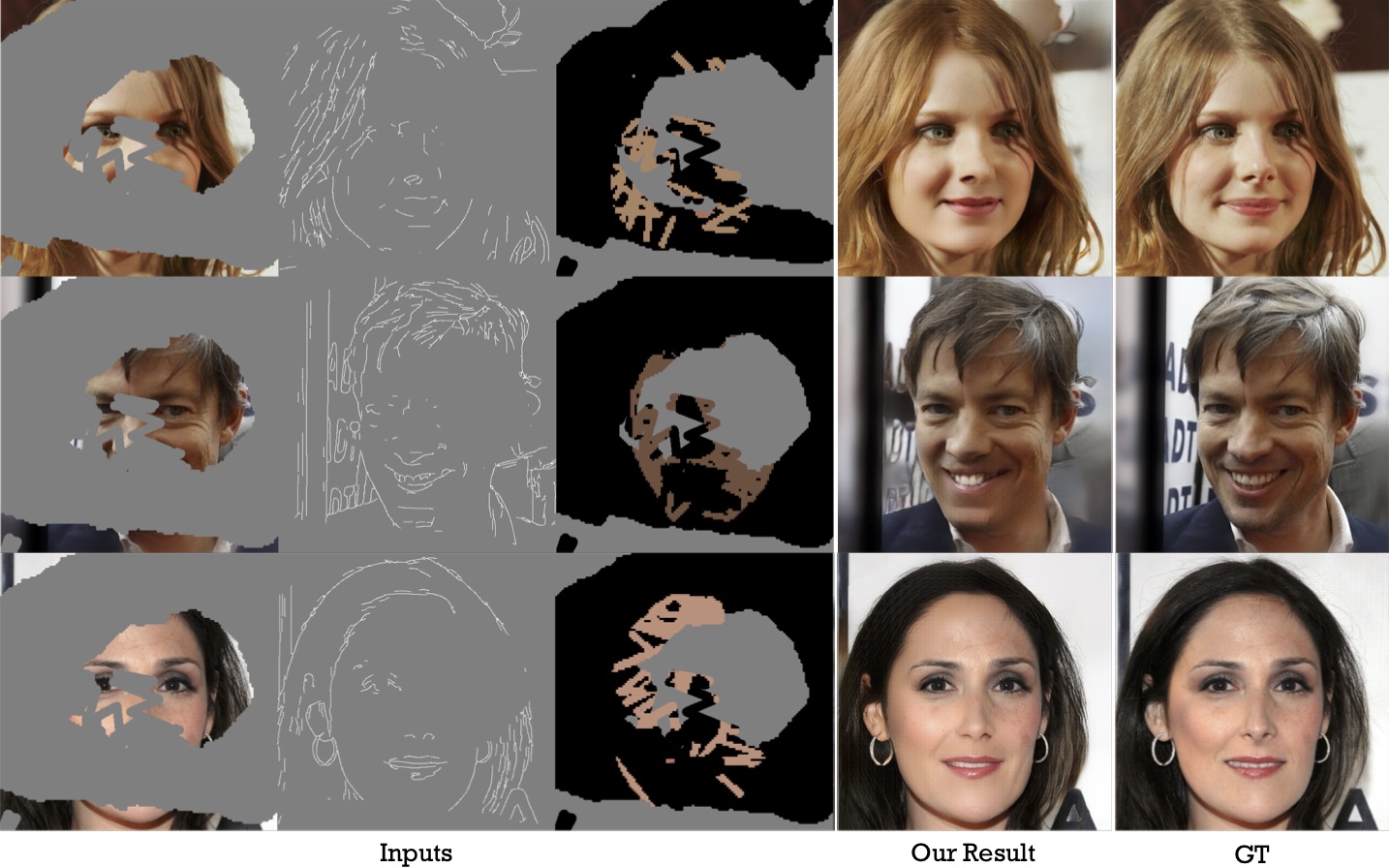}
\end{center}
   \caption{Our results about face restoration. Our system can restore the face satisfactorily if given enough input information even if a lot of regions are erased.}
\label{fig:8}
\vspace{0.5cm}
\end{figure*}

\begin{figure*}
\begin{center}
   \includegraphics[width=1.0\linewidth]{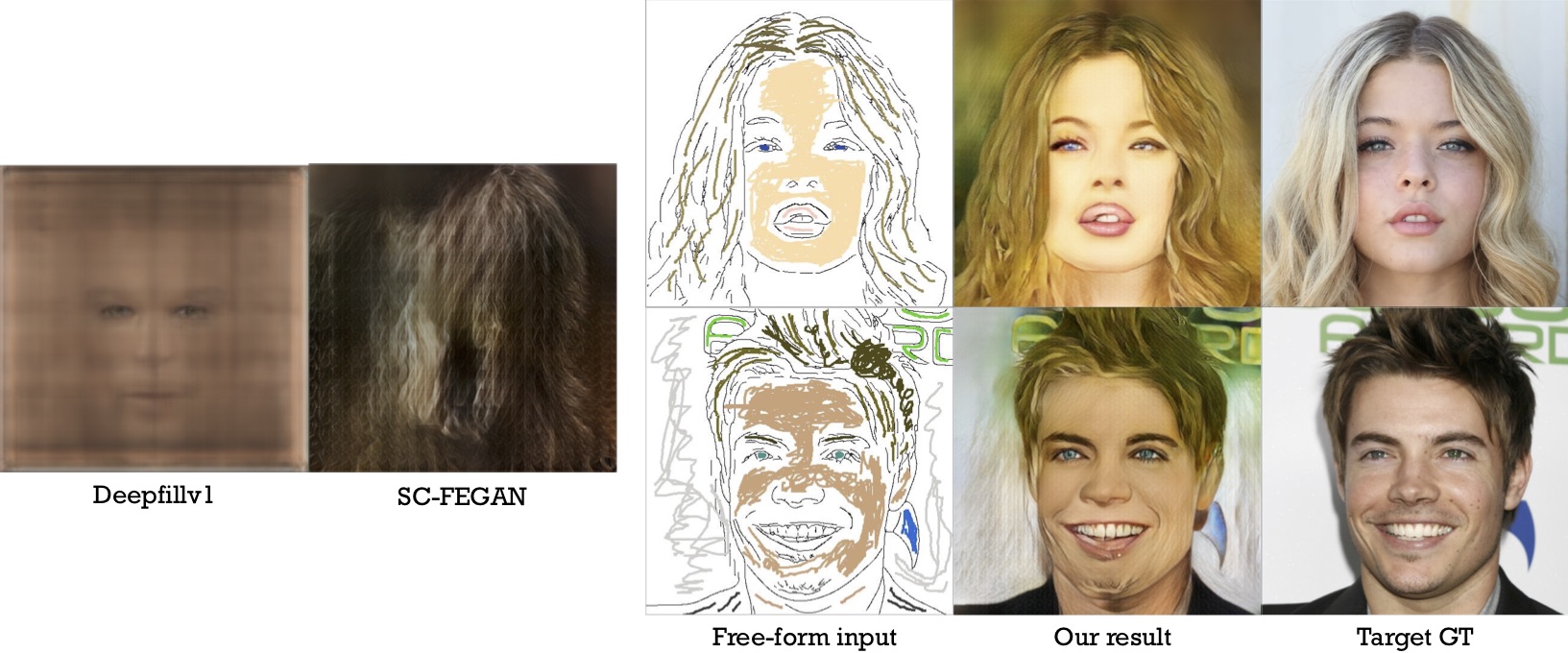}
\end{center}
   \caption{Our results about total restoration. On the left, it shows that results of Deepfillv1~\cite{yu2018generative} and SC-FEGAN about totally erased image. On the right, it shows that SC-FEGAN can be works like translation. It can generates face image only with sketch and color input.}
\label{fig:10}
\end{figure*}

Figure~\ref{fig:7} shows various results with sketch and color inputs.
It shows that our system allows users to intuitively edit the face image features such as hair style, face shape, eyes, mouth \textit{etc}.
Even if the entire hair region is erased, our system is capable of generating an appropriate result once it is provided with a user sketch.
Users can intuitively edit the images with sketches and colors, while the network tolerates small drawing error.
The user can modify the face image intuitively through sketch and color input to obtain a realistic synthetic image that reflects shadows and shapes in detail.
Figure~\ref{fig:8} shows some results for the validation dataset, which shows that even though the user has a lot of modifications, the user can get a high quality composite image with enough user input.
In addition, in order to check the reliance on the dataset the network learned, we tried to input the erased image of all areas.
Compared with Deepfillv1~\cite{yu2018generative}, Deepfillv1 generates faint image of face but our SC-FEGAN generates faint image of hair (see Figure~\ref{fig:10}).
It means that without additional information, like sketch and color, the shape and position of elements of face have a certain dependent value. Therefore, it is only necessary to provide additional information to restore the image in the desired direction.
In addition, our SC-FEGAN can generates face image with only sketch and color free-form input even if the input image is erased totally (see Figure~\ref{fig:10}).
\subsection{Interesting results}
\begin{figure*}
\begin{center}
   \includegraphics[width=1.0\linewidth]{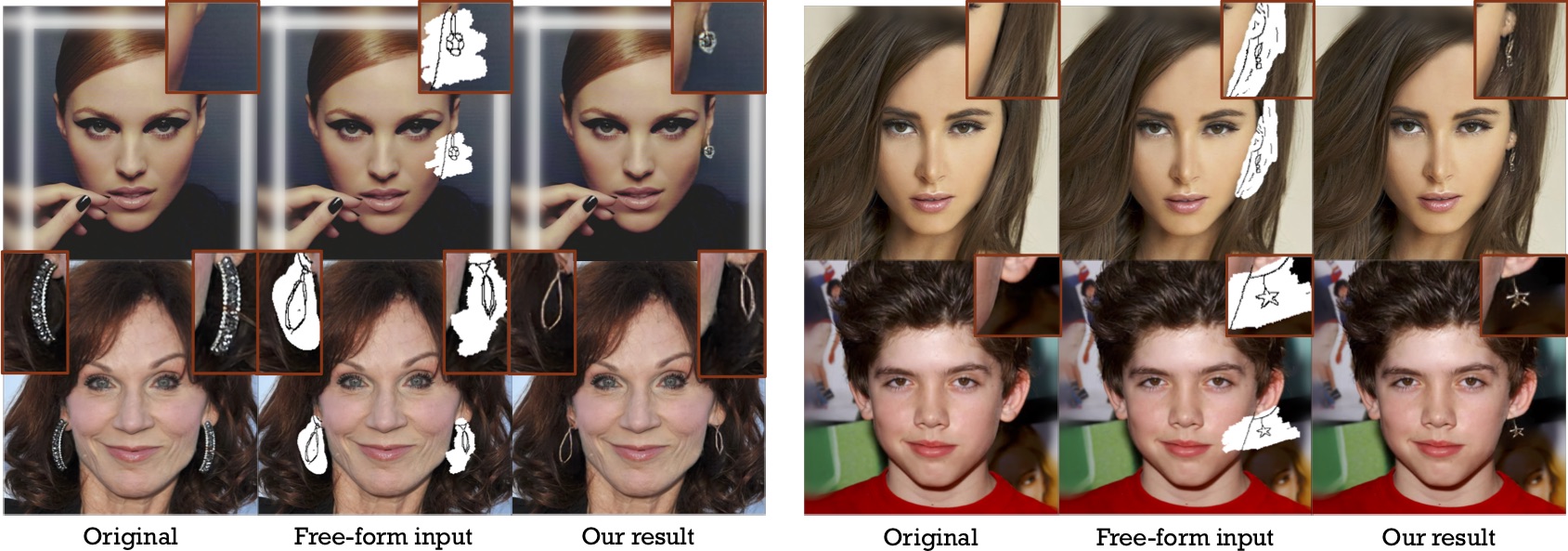}
\end{center}
   \caption{Our special results about edit earrings.}
\label{fig:9}
\end{figure*}

The image results generated by the GAN often show high dependencies on the training data sets.
Deepfillv2~\cite{yu2018free} used same dataset CelebA-HQ, but used only landmark to make sketch dataset.
In Faceshop~\cite{portenier2018faceshop}, the AutoTrace~\cite{autotrace} erased small details in the images of the dataset.
In our study, we applied HED to all area, and by scheduling it to extend the masking area, we were able to obtain special results that produces facial image along with earrings.
Figure~\ref{fig:9} shows selection of such interesting results.
These examples demonstrate that our network is capable of learning small details and generate reasonable results even for small input.

\section{Conclusions}
In this paper we present a novel image editing system for free-form  masks, sketches, colors inputs which is based on an end-to-end trainable generative network with a novel GAN loss.
We showed that our network architecture and loss functions significantly improve inpainting results in comparison with other studies.
We trained our system on high resolution imagery based on the celebA-HQ dataset and show a variety of successful and realistic editing results in many cases.
We have shown that our system is excellent at modifying and restoring large regions in one pass, and it produces high quality and realistic results while requiring minimal efforts from the users.

{\small
\bibliographystyle{ieee}
\bibliography{egbib}
}

\end{document}